\documentclass{article} 
\usepackage{iclr2025_conference,times}

\usepackage{amsmath,amsfonts,bm}

\def\eqref#1{equation~\ref{#1}}

\def\1{\bm{1}}

\DeclareMathAlphabet{\mathsfit}{\encodingdefault}{\sfdefault}{m}{sl}
\SetMathAlphabet{\mathsfit}{bold}{\encodingdefault}{\sfdefault}{bx}{n}

\usepackage{xcolor}         
\usepackage{url}
\usepackage{enumitem}
\usepackage{algorithm}
\usepackage{algpseudocode}
\usepackage{graphicx}
\setlist[itemize]{noitemsep, topsep=0pt}
\usepackage{tikz}
\usepackage[perpage]{footmisc}
\usepackage{authblk}

\usepackage{xspace}
\usepackage{wrapfig}
\newcommand{\method}{{MetaInf}\xspace}

\newtheorem{problem}{Problem}

\definecolor{darkgreen}{RGB}{60,179,113}
\newcommand{\tikzcircle}[2][red,fill=red]{\tikz[baseline=-0.5ex]\draw[#1,radius=#2] (0,0) circle ;}%

\definecolor{lightgreen}{HTML}{39b54a}
\usepackage[pagebackref=false, colorlinks=true, citecolor=lightgreen]{hyperref}

\title{Meta-Learning for Speeding Up Large Model Inference in Decentralized Environments}

\author{%
    Yuzhe Yang, Yipeng Du, Ahmad Farhan, Claudio Angione,
    Yue Zhao, Harry Yang, \\[0.2em]
    Fielding Johnston,
    James Buban,
    Patrick Colangelo
}

\affil[]{Nesa Research}
\affil[]{\texttt{research@nesa.ai}}

\iclrfinalcopy 
\begin{document}

\begin{center}
    \maketitle
\end{center}

\begin{abstract}
The deployment of large-scale models, such as large language models (LLMs) and sophisticated image generation systems, incurs substantial costs due to their computational demands. 
To mitigate these costs and address challenges related to scalability and data security, there is a growing shift towards decentralized systems for deploying such models. 
In these decentralized environments, efficient inference acceleration becomes crucial to manage computational resources effectively and enhance system responsiveness.
In this work, we address the challenge of selecting optimal acceleration methods in decentralized systems by introducing a meta-learning-based framework. This framework automates the selection process by learning from historical performance data of various acceleration techniques across different tasks. Unlike traditional methods that rely on random selection or expert intuition, our approach systematically identifies the best acceleration strategies based on the specific characteristics of each task. 
We demonstrate that our meta-learning framework not only streamlines the decision-making process but also consistently outperforms conventional methods in terms of efficiency and performance.
Our results highlight the potential of meta-learning to revolutionize inference acceleration in decentralized AI systems, offering a path towards more democratic and economically feasible artificial intelligence solutions.\footnote{Work In Progress.}
\end{abstract}


\section{Introduction}

The advancement of large-scale models such as large language models (LLMs) and sophisticated image generation systems has dramatically increased computational demands, necessitating significant innovation in deployment architectures \citep{brown2020language, ramesh2022hierarchical}. Traditional centralized systems, while powerful, encounter critical limitations in terms of scalability, data security, and operational costs \citep{li2022survey}. These limitations have spurred interest in decentralized architectures that distribute computational tasks across multiple nodes to enhance efficiency, reduce latency, and improve data privacy \citep{belotti2019vademecum}.

Decentralized systems, by their nature, facilitate a more robust approach to managing data and computational resources, offering an attractive solution for deploying computationally intensive models \citep{xu2019geeps}. Such architectures are crucial for supporting real-time processing and high availability across varied geographical locations without compromising on performance and security \citep{wang2019edge}.



\begin{figure}[!th]
  \centering
  \includegraphics[width=1.0\columnwidth]{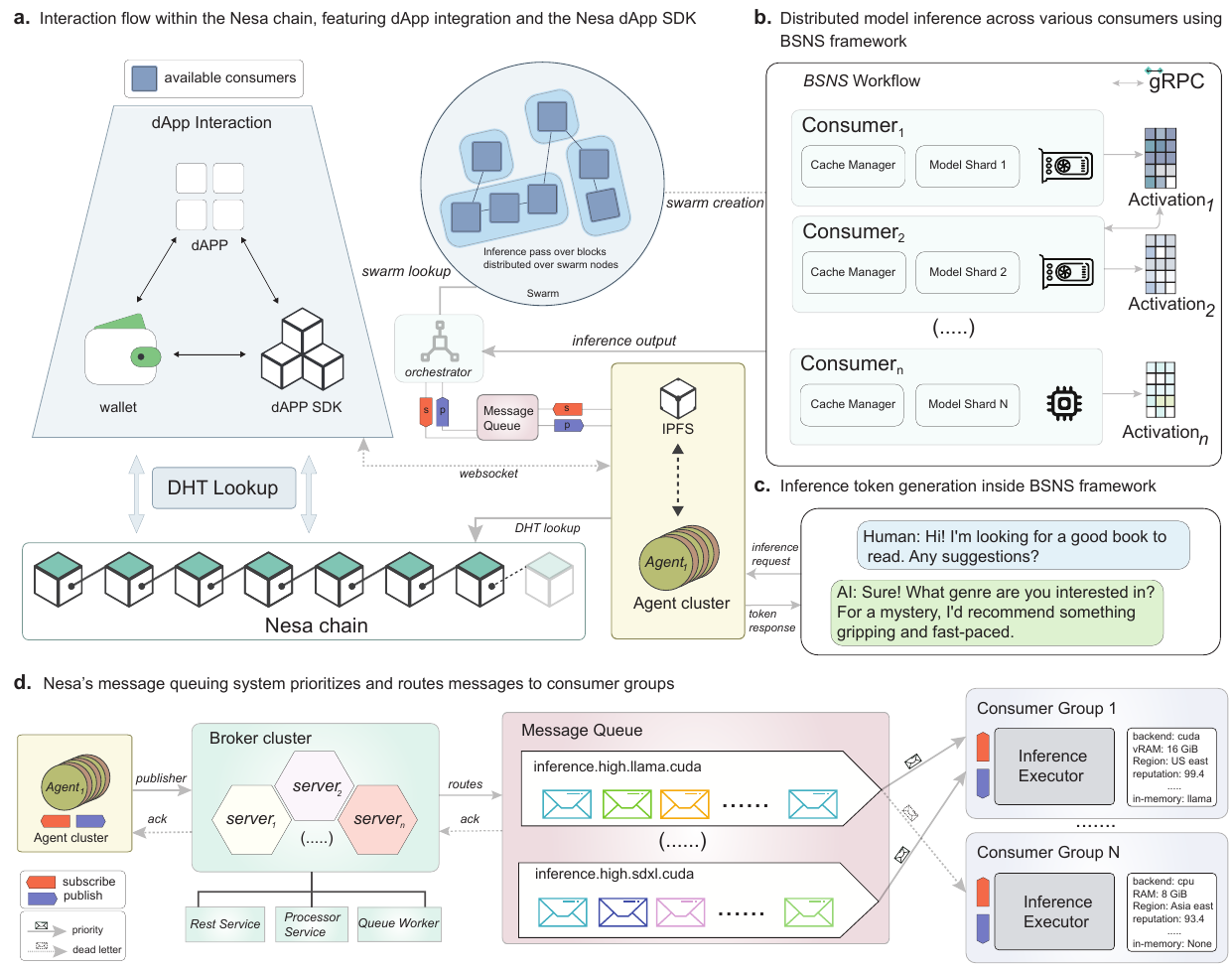}
  \caption{\textbf{BSNS framework overview: a.} Interaction flow within the Nesa chain highlights the sequence from dApp interaction to final output. It begins with a query through a dApp and wallet, integrates with the dApp SDK, and involves a DHT lookup across the Nesa chain for blockchain transactions. \textbf{b.} Shows BSNS framework's role in distributed model inference, where different consumers, each managing a model shard, process an inference request in parallel. Blocks are distributed among swarm nodes, with the activations sequentially processed and managed through gRPC communication. \textbf{c.} For an LLM text generation query, the agent reads the completed generate response from the queue and performs a DHT lookup to validate the transaction against the Nesa chain, and sequentially delivers the response back to the dApp. \textbf{d.} Shows the message queuing system within Nesa's architecture, where an agent cluster publishes requests to a broker cluster. The broker prioritizes and routes these requests to appropriate consumer groups based on resource allocation, reputation scores and model requirements.}
  \label{fig:bsns_architecture}
\end{figure}

\subsection{Nesa's system overview}

To enable AI democratization, Nesa proposed a model-agnostic hybrid sharding approach named BSNS, layered onto a hardware-based and software-based privacy co-optimization protocol \cite{angione2024model}. The idea is to distribute the computational load of an inference request across multiple nodes on a decentralized inference network. The protocol also incorporates homomorphic encryption, therefore delivering privacy-preserving data processing. As a result, anyone can perform inference queries even with limited local computing power (e.g. no GPU). 

At the core of our protocol, we developed a dynamic routing mechanism. More specifically, the effective routing of data and tasks is crucial in a distributed blockchain network designed for inference tasks. The routing decision specifies a subset of consumers that are chosen to serve as the computational resources that run arbitrary neural network blocks. The full set of blocks, which constitute any given model, is therefore distributed and executed across the network. \begin{equation}
\mathcal{S}: (A_i)_{i \in \{1, \ldots, p\}}
\end{equation}

where \(\mathcal{S}\) represents the sequence of consumers, with each \(A_i\) denoting the \(i\)-th shard managed by the \(i\)-th consumer (Fig.~\ref{fig:bsns_architecture}a). This setup reduces latency and bandwidth needed for real-time inference. Importantly, our routing mechanism is dynamically updated based on each consumer's features, including hardware-based and reputation-based metrics.

For distributed inference within the network, Nesa uses a message queuing system as shown in Figure~\ref{fig:bsns_architecture}d. Inference tasks from an agent cluster are received by a broker cluster, which prioritizes them before routing to the appropriate consumer groups. Tasks are queued based on computational demand and response time requirements. Consumer groups, distinguished by their hardware capabilities and geographical distribution, execute these tasks. This prioritization and routing are informed by a reputation heuristic that ensures tasks are handled by nodes with a history of reliability and optimizes resource use and minimizing latency. 

\textbf{Challenges of Fast Inference in Decentralized Systems}.
However, managing inference acceleration within these decentralized systems presents a unique set of challenges, primarily due to the diverse and often constrained computational resources available at different nodes \citep{deng2020model}. To address these challenges effectively, there is a pressing need for adaptive strategies that can dynamically select the most appropriate inference acceleration methods based on the specific characteristics of the task and the underlying system constraints \citep{he2020group}.

\textbf{Our Solution:}
In response, we extend the BSNS architecture by introducing a novel meta-learning framework, \method, designed to optimize the selection of inference acceleration methods within decentralized systems. This framework leverages historical performance data to train a meta-learner capable of predicting the most effective acceleration strategies under various operational scenarios, thereby enhancing the computational efficiency of large model deployments and making high-performance AI more accessible and cost-effective \citep{chen2019learning}. \method streamlines the inference process and shows substantial improvements over traditional methods that rely on either random selection or solely expert-driven strategies \citep{wang2021survey}.

\textbf{Contributions:}
Our work makes several contributions to the field of AI model deployment:
\begin{enumerate}[leftmargin=*]
    \item We present the first comprehensive framework for fast inference that effectively addresses the unique challenges of large model deployments in decentralized systems.
    \item We introduce a novel meta-learning-based approach for selecting inference acceleration methods, optimizing computational resources across heterogeneous environments.
    \item We demonstrate the effectiveness of our framework through extensive experiments, showing significant improvements in efficiency and cost-effectiveness over traditional inference acceleration methods.
\end{enumerate}

The following sections will first discuss the fast inference approaches in \S \ref{sec:acceleration-methods}, the challenge of extending to decentralized systems (\S \ref{sec:distributed-setting}), and our meta-learning-based selection framework (\S \ref{sec:meta-learner})


\section{A Closer Look at Fast Inference}
\label{sec:acceleration-methods}

Fast inference has become a crucial factor in the deployment of large models and complex AI systems \citep{brown2020language}. In many real-world scenarios, models need to process user queries and produce results in a fraction of a second \citep{rajbhandari2020zero}. This requirement poses challenges in both centralized and decentralized environments, particularly as model sizes and the demand for real-time processing grow. In this section, we first analyze state-of-the-art fast inference methods in \textit{centralized} systems, benchmarking their performance across a variety of tasks, models, and hardware settings.

\subsection{Language Models}

Autoregressive (large) language models have seen significant advancements in fast inference \citep{kwon2023efficient, leviathan2023fast}. Many approaches are aimed at reducing latency and maximizing throughput without compromising model accuracy. We identify and summarize several key techniques, evaluating their efficacy and trade-offs.

\paragraph{Large Batch Size.}
One of the most straightforward methods for optimizing inference is batching, where multiple requests are grouped into a single batch and processed simultaneously. GPUs, as highly parallel processors, operate more efficiently when handling large batches, which leads to a significant increase in throughput. The main advantage of large batch sizes is the improved GPU utilization, especially when multiple inference requests are made at once.
This method has been successfully applied in various production systems to scale LLMs for real-time use cases \citep{shoeybi2019megatron, rajbhandari2020zero, henighan2020scaling}.

\paragraph{Continuous Batching.}
The static batching strategy, which uses a fixed batch size, while effective for parallel processing, often introduces latency, as the system waits for sufficient requests to form a batch. To reduce this inefficiency, \textit{dynamic batching} was the first improvement in this area, where batches are processed either when a certain size is reached or after a set time window. 
\textit{Continuous batching} extends this concept by dynamically integrating incoming requests as they arrive, allowing the GPU to maintain high utilization with minimal idle periods \citep{yu2022orca,kwon2023efficient}. By batching requests at the token level and interleaving prefill and decoding stages, continuous batching enables seamless incorporation of new tokens into the processing flow \citep{yu2022orca}. This approach ensures that the model can handle real-time requests more efficiently, improving overall throughput and reducing latency by interleaving the completion of earlier requests with the arrival of new ones \citep{rajbhandari2020zero, shoeybi2019megatron}.

\paragraph{Prefilling.}
In autoregressive language models, the key-value (KV) cache is crucial for efficient generation, as it stores intermediate states of input tokens. Prefilling refers to the precomputation of the KV cache for input tokens before the model begins generating output \citep{pope2023efficiently}. This precomputation reduces the computational burden during generation, especially for long input sequences, where input tokens are processed significantly faster \citep{kwon2023efficient}. By optimizing the prefill phase, the overall latency of the model can be minimized, leading to substantial performance gains, particularly in real-time applications \citep{brown2020language, radford2019language}.

\paragraph{Prompt Caching.}
Large language models frequently encounter repeated prefixes in prompts, such as ``\texttt{you are a smart AI agent, ...}'', especially in system prompts that are reused across multiple requests. Prompt caching avoids recalculating the outputs for these repeated segments by storing prefilled prompts and reusing them when the same prefix appears. This caching mechanism eliminates redundant computations, enabling the model to skip over already processed inputs and focus solely on the new portions of the request.
This is particularly beneficial in real-time applications where similar queries are frequently submitted, allowing for faster response times and more efficient use of computational resources \citep{rajbhandari2020zero, brown2020language}.

\paragraph{Speculative Decoding.}
Speculative decoding aims to accelerate the inference process of autoregressive models by leveraging faster, smaller models to predict likely token candidates before invoking a larger, more computationally expensive model \citep{leviathan2023fast}. The core idea is to use a lightweight auxiliary model to generate multiple possible token sequences in parallel, allowing the larger model to confirm or refine these predictions rather than generating each token sequentially from scratch. This approach takes advantage of the observation that many tokens, especially in common phrases or predictable contexts, can be accurately predicted by simpler models, thus offloading a significant portion of the computational burden.

Recent advancements in speculative decoding have further refined the technique by improving speculative model \citep{kwon2023efficient, cai2024medusa} and better feature uncertainty \citep{li2024eagle}.
In this paper, we focus on speculative decoding methods through the vLLM framework \citep{kwon2023efficient}.

\paragraph{Quantization.}
Quantization is widely-used in deep learning to reduce the computational and memory demands of large models by lowering the precision of the model’s parameters and activations. Instead of using 32-bit or 16-bit floating-point numbers (\texttt{fp32} or \texttt{fp16}), quantization converts these values to lower-precision formats such as 8-bit integers (\texttt{int8}) or even smaller, which drastically reduces the memory footprint and computational cost \citep{kwon2023efficient}. The primary motivation for quantization is that many neural network operations can still achieve competitive performance using lower precision, without significant degradation in model accuracy. This has made quantization an attractive method for deploying large models like LLMs and image generation systems in resource-constrained environments such as mobile devices or edge computing platforms \citep{dettmers2022llm, rajbhandari2020zero}.

In this paper, we explore the tradeoff in quantization between model efficiency and accuracy.

\subsection{Text-to-Image Models}
In addition to autoregressive models such as LLMs, serving text-to-image (T2I) generation models (e.g., diffusion models like Stable Diffusion \citep{rombach2022high}) presents its own unique challenges for optimizing inference speed. Improving latency and generation speed is essential to enable real-time interaction, fast response, and seamless user experiences \citep{dhariwal2021diffusion}. Particularly in resource-constrained or real-time environments, it becomes critical to identify and implement key techniques for accelerating T2I model inference.

\paragraph{Distilled Models.}
Neural network distillation offers a more efficient alternative to full-sized models by reducing the number of residual and attention blocks \citep{hinton2015distilling}. This distillation process results in a smaller, faster model that maintains image quality while significantly reducing memory usage and inference time. For example, a distilled version of Stable Diffusion can achieve up to a $\sim$1.5X speedup, and when paired with a distilled autoencoder, this performance boost can reach up to $\sim$1.6X compared to baseline models \citep{meng2023distillation, ho2020denoising}.

\paragraph{Reduced Precision.}
Utilizing reduced numerical precision formats, such as \texttt{bfloat16}, can significantly improve inference latency without impacting the overall quality of image generation \citep{ansel2024pytorch}. Modern GPUs feature specialized cores that are optimized for lower-precision computations, allowing them to process operations faster than with higher precision formats like \texttt{float32} or even \texttt{float16}. The robustness of \texttt{bfloat16} compared to \texttt{float16} makes it particularly suitable for scenarios where quantization is applied, as it helps maintain model accuracy despite the reduced precision \citep{gupta2015deep}.

\paragraph{Scaled Dot Product Attention.}
Scaled Dot Product Attention (SDPA) optimizes the computationally intensive attention-related operations that are central to transformer-based models like the UNet and Variational Autoencoder (VAE) components used in diffusion models \citep{ansel2024pytorch}. By implementing SDPA, PyTorch 2.0 introduces more efficient kernels that reduce the overhead of attention calculations, speeding up inference while maintaining accuracy \citep{ansel2024pytorch}. This optimization addresses a common bottleneck in high-dimensional image generation tasks, particularly when processing complex attention maps.

\paragraph{Combined Projections for Attention Operations.}
In transformer-based architectures, including those used in T2I models, the self-attention mechanism typically projects the input into three separate subspaces (queries, keys, and values) using distinct matrices. By combining these three projections into a single matrix and performing the operation simultaneously, the model can benefit from more efficient matrix multiplications \citep{child2019generating}. This larger matrix multiplication allows for better utilization of GPU hardware, particularly in scenarios that benefit from quantization, enhancing both speed and memory efficiency.

\paragraph{Dynamic Quantization.}
Dynamic INT8 quantization reduces the precision of specific layers in the UNet and VAE models, leading to faster inference, particularly for larger matrix multiplications. Although quantization can introduce overhead due to data type conversions, selectively applying it to layers where it delivers the most benefit, such as pointwise convolution layers converted into linear layers, can optimize the overall performance without a significant loss in image quality \citep{nagel2021white, han2016deep}.

\subsection{Implementation}

\begin{figure}[!t]
  \centering
  \includegraphics[width=0.99\columnwidth]{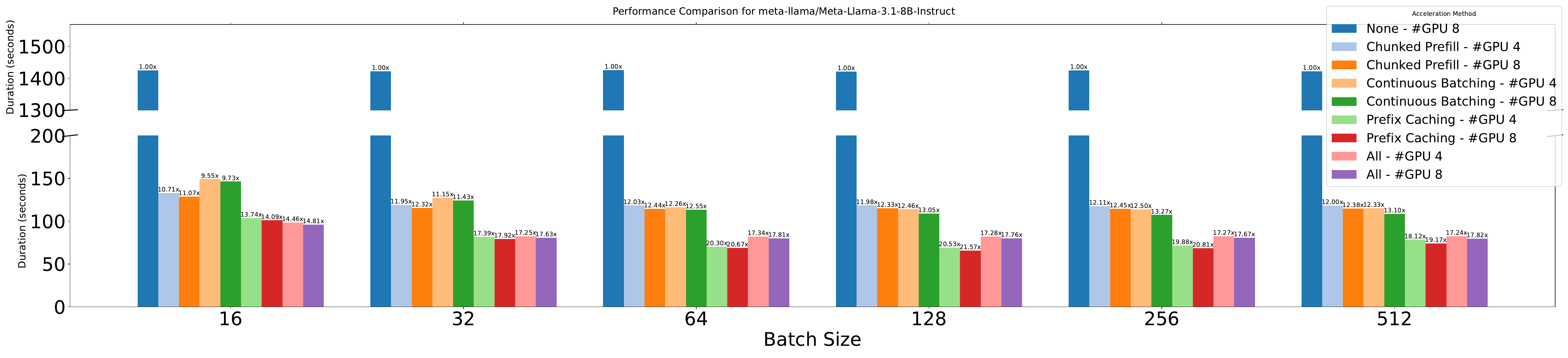}
  \caption{Motivating example: Performance comparison of fast inference methods on Llama 3.1 8B.}
  \label{fig:motivation}
\end{figure}

We implement and integrate the aforementioned methods into your system, and carry out a comprehensive study to understand the performance of different fast inference methods in heterogeneous setup (Figure \ref{fig:motivation}). 

\paragraph{Experimental Setup.}
We conducted our experiments using the ShareGPT dataset \citep{kwon2023efficient} for throughput testing. NVIDIA L4 GPUs were used for testing. Tensor parallelism was varied across different number of GPU configurations.
The models tested include \textit{Meta-Llama-3.1-8B-Instruct}, \textit{Microsoft Phi-2}, and \textit{Mistral-7B-v0.1}, with all outputs limited to 10 tokens. We evaluated \textit{continuous batching, prefix caching, chunked prefill, and all enabled} of these methods, but speculative decoding and quantization were excluded due to hardware and time limitations.


We summarize the key findings as follows:

\begin{itemize}[leftmargin=*]
    \item \textbf{Significant improvement with fast inference methods.} Across all tested models, the introduction of fast inference methods such as continuous batching and prefix caching resulted in a marked increase in throughput. In particular, we observed that continuous batching consistently reduced latency by up to \textbf{13X} compared to the baseline, and prefix caching can reach up to \textbf{20X} inference speed gains.
    \item \textbf{Different gains from different methods.} The individual contributions of different fast inference methods varied significantly. Prefix caching yielded the most substantial speedup for models (Figure \ref{fig:motivation}). Chunked prefill, while less impactful than batching or prefix caching, still provided a noticeable improvement. 
    \item \textbf{The effect of different numbers of GPUs.} We also examined the performance impact of varying the number of GPUs with tensor parallelism. As expected, increasing the number of GPUs led to higher throughput, but the relative gains diminished beyond a certain point. For example, moving from 4 GPUs to 8 GPUs resulted in only a \textbf{5\%} gain. This diminishing return suggests that communication overhead and memory bandwidth limitations become bottlenecks when scaling the number of GPUs and increasing the number of pipelines when serving might be a better use of additional hardware.
\end{itemize}

\section{Fast Inference in Heterogeneous Systems}
\label{sec:distributed-setting}

Building on the state-of-the-art techniques evaluated for improving inference speed in \textit{centralized} architectures, we extend our analysis to explore how these methods can be adapted to \textit{decentralized} systems. In distributed settings, achieving both scalability and efficiency presents unique challenges due to heterogeneous hardware configurations and variable system conditions.

\paragraph{Conceptualization.}
Effective AI deployment in decentralized systems requires careful consideration of heterogeneous hardware resources, network latency, and dynamic load balancing \citep{borzunov2024distributed}. Unlike in centralized architectures, where inference tasks are managed by a single or a small number of powerful machines, decentralized settings must accommodate various hardware setups. This includes devices with differing computational power, memory constraints, and communication overheads. The challenge of distributing workloads across this diverse infrastructure is non-trivial and requires dynamic scheduling based on the available hardware, current system throughput, and overall network status.

\paragraph{Observations from the Motivating Example.}
As illustrated in Figure \ref{fig:motivation}, our experiments evaluated the effectiveness of applying fast inference techniques, including continuous batching and prefix caching, under varying conditions. In environments with consistent hardware capabilities, these methods demonstrated significant improvements in inference speed. However, when varying the batch sizes, the performance gains became less predictable. Specifically, for smaller batch sizes, combining all techniques (``All'') yielded the best results. In contrast, with larger batch sizes, the performance benefits diminished, and single-method strategies were more effective. Although this is a simplified setup, it highlights a key observation: in heterogeneous systems, performance varies based on the distribution strategy and the real-time adaptation of workload scheduling. 

\paragraph{Interpretation.}
Our results highlight the importance of flexibility in distributed AI systems and the need for \textbf{\textit{adaptive}} frameworks to handle the inherent variability of decentralized environments.
Inspired by these observations, we propose a \textit{learnable} meta-scheduler, which dynamically predicts the optimal methods and hardware configurations to utilize, based on the input data and current system status. The details of this adaptive meta-scheduler are discussed further in Section \ref{sec:meta-learner}, where we outline its architecture and implementation for fast inference across distributed systems.

\section{Meta-Learning for Inference Acceleration Optimization}
\label{sec:meta-learner}

\begin{figure}[!th]
  \centering
  \includegraphics[width=0.98\columnwidth]{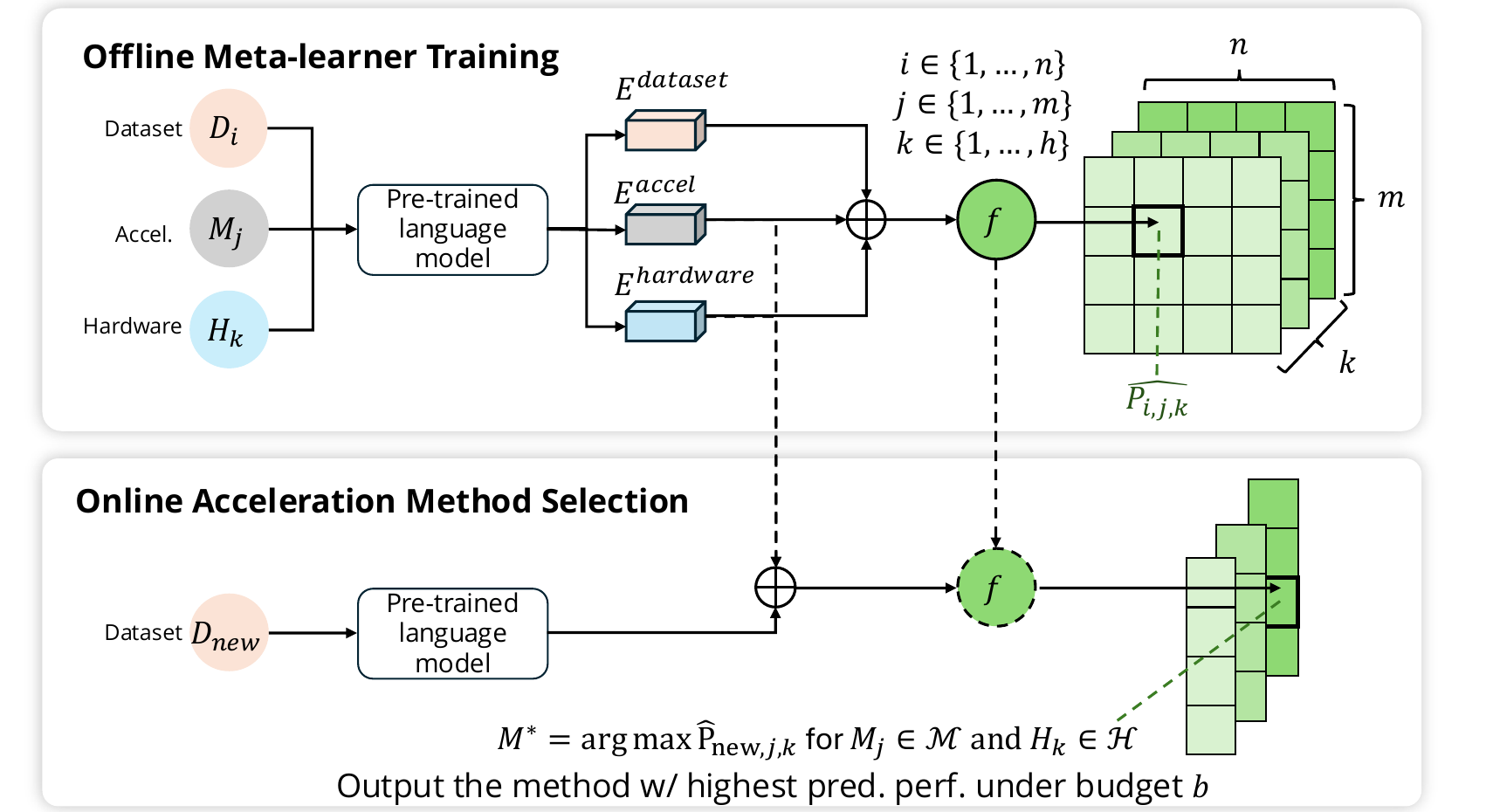}
  \caption{\method overview (\S \ref{subsec:overview}); offline meta-training phase is shown on the top (\S \ref{subsec:meta-train})--the key is to train a meta performance predictor $f$ (denoted in \tikzcircle[darkgreen, fill=darkgreen]{3pt}) to map language embeddings of the datasets and models to their performance $\mathbf{P}$; the online model selection (\S \ref{subsec:model-selection}) is shown at the bottom by transferring the meta-predictor $f$ to predict the test data paired with acceleration methods and hardware settings for selection.
  }
  \label{fig:pipeline}
\end{figure}

\subsection{Problem Statement and Framework Overview}
\label{subsec:overview}

Given a new model, dataset, and hardware environment for distributed inference, the goal is to select the best acceleration method from a heterogeneous set of techniques such as quantization, model compression, and parallelism, \textit{without} requiring extensive empirical evaluations. This selection process must account for different hardware platforms' unique computational constraints and capabilities.

In this work, we leverage meta-learning to transfer acceleration performance information from prior experiences to the new inference task. Meta-learning, often referred to as ``learning to learn," is a technique where an algorithm learns from a collection of historical or meta tasks and uses this experience to perform well on new, unseen tasks. The rationale is that an acceleration method that performed well in a similar historical context is likely to outperform other methods on a new dataset, model configuration, and hardware setup. This approach is particularly useful when immediate evaluation is infeasible or costly due to computational constraints or the need for rapid deployment.

The proposed meta-learner, \method, is designed to optimize inference acceleration across diverse computing environments by learning from historical data. It incorporates several key components:

\begin{itemize}[leftmargin=*]
    \item A detailed database of historical inference tasks, $\mathcal{D}_\text{train} = \{D_{1}, \ldots, D_{n}\}$, provides a broad learning base from which \method can draw insights. Each task in this database is defined by a unique input to the machine learning models, e.g., prompts for LLMs.

    \item A catalog of hardware environments, $\mathcal{H} = \{H_1, \ldots, H_h\}$, aids in understanding how different configurations influence the efficacy and cost-efficiency of acceleration methods.

    \item Performance metrics, $\mathbf{P}$, associated with a set of predefined acceleration methods, $\mathcal{M} = \{M_1, \ldots, M_m\}$, are measured under various conditions to form a comprehensive performance matrix $\mathbf{P} \in \mathbb{R}^{n \times m \times h}$. Each element $\mathbf{P}_{i,j,k}$ provides insights into the performance of method $M_j$ for task $D_i$ on hardware $H_k$.
\end{itemize}

    
    

Our objective is to identify the best acceleration method $M \in \mathcal{M}$ for new tasks, considering not only performance but also cost constraints defined by the budget \( b \). The selected method must ensure that the cost of deployment on hardware \( H_k \), defined as the product of hardware cost and predicted runtime, does not exceed \( b \).

\begin{problem}[Cost-Constrained Inference Acceleration Method Selection]
Given a new task $D_\text{new} = \{\mathbf{X}_{\text{train}}^\text{new}, \mathbf{X}_{\text{test}}^\text{new}, \mathbf{H}_\text{new}\}$ and a cost budget \( b \), select an acceleration method $M \in \mathcal{M}$ such that the cost of using $M$ on hardware $H_k$ for $D_\text{new}$ is minimized while not exceeding \( b \).
\end{problem}

Our \method framework, detailed in the subsequent sections, consists of two phases: (\textit{i}) offline meta-training on $\mathcal{D}_\text{train}$ to learn historical performance mappings, and (\textit{ii}) online method selection that applies learned insights to effectively and economically deploy new tasks. Figure~\ref{fig:pipeline} provides an overview of these processes, and further details are discussed in Sections \ref{subsec:meta-train} and \ref{subsec:model-selection}.

\subsection{Offline Meta-Training}
\label{subsec:meta-train}

During the offline training phase, we generate embeddings that represent the characteristics of the datasets, the different acceleration methods, and the hardware environments used historically. This comprehensive approach is crucial for training the latent mappings that correlate these embeddings to the observed performance metrics, denoted as $\mathbf{P}$.

\begin{equation} \label{eq:1}
{f : E_{i}^\text{data}, E_{j}^\text{model}, E_{k}^\text{hardware}}
\rightarrow \mathbf{P}_{i,j,k}, \quad i \in \{1, \dots, n\}, \quad j \in \{1, \dots, m\}, \quad k \in \{1, \dots, h\}
\end{equation}

The embeddings for the datasets $\mathcal{D}_\text{train}$, methods $\mathcal{M}$, and hardware configurations $\mathcal{H}$ are generated to capture the essential characteristics that influence the efficiency and efficacy of the acceleration methods under various computational and environmental constraints. These embeddings, referred to as $E^\text{data}$, $E^\text{model}$, and $E^\text{hardware}$ respectively, are derived from a comprehensive analysis of each dataset’s, method's, and hardware's features.

To train the meta-learner that predicts the best-performing acceleration method for new, unseen tasks, we employ a regression-based meta-predictor. Inputs to this predictor are the embeddings of the datasets, models, and hardware, which encapsulate critical performance-influencing factors:

\textbf{Dataset Embedding} ($E^\text{data}$): Generated by encoding characteristics such as data volume, complexity, distribution, and typical processing requirements crucial for selecting an appropriate acceleration method.

\textbf{Method Embedding} ($E^\text{model}$): Created by encoding features such as the method's impact on computational overhead, energy efficiency, and potential accuracy trade-offs, including method-specific parameters like degree of quantization, levels of compression, or the architecture required for parallel execution.

\textbf{Hardware Embedding} ($E^\text{hardware}$): Encodes the specific hardware capabilities and limitations, such as processor type, memory availability, and power consumption, which can significantly affect the performance of acceleration methods.

The meta-predictor, denoted as $f$, is trained using a dataset of historical performance metrics. This dataset comprises various combinations of dataset characteristics, method efficacies, and hardware settings, represented by the embeddings. We employ a model such as XGBoost \citep{chen2016xgboost} due to its effectiveness in handling diverse and high-dimensional data, as well as its capability for feature importance evaluation, which is vital for understanding which characteristics most significantly impact performance.

\textbf{Objective}: The primary goal during this training phase is to establish a reliable mapping from $(E_{i}^\text{data}, E_{j}^\text{model}, E_{k}^\text{hardware})$ to $\mathbf{P}_{i,j,k}$, where $\mathbf{P}_{i,j,k}$ might include metrics like runtime efficiency, energy usage, and accuracy. This mapping allows us to predict which acceleration method will most likely yield the best performance for a given new task characterized by $D_\text{new}$.

The steps of the offline meta-training phase, including the generation of embeddings and training of the meta-predictor, are visualized in Figure~\ref{fig:pipeline}, 
which is detailed further in the subsequent sections and Algo. \ref{alg:offline}. 

\begin{algorithm}
\caption{Offline Meta-Learner Training for Inference Acceleration}
\label{alg:offline}
\renewcommand{\algorithmicrequire}{\textbf{Input:} Meta-train database \( \mathcal{D}_{\text{train}} \), model set \( \mathcal{M} \), hardware environments \( \mathcal{H} \)}
\renewcommand{\algorithmicensure}{\textbf{Output:} Meta-learner $f$ for acceleration method selection considering hardware adaptations}
\begin{algorithmic}[1]
\Require $ $
\Ensure $ $
\State Train and evaluate $\mathcal{M}$ across $\mathcal{H}$ on $\mathcal{D}_{\text{train}}$ to get performance tensor $\mathbf{P}$
\For {$i \in \{1, \dots, n\}$}
    \State Extract data embedding $E^\text{data}_{i} = \psi(D_{i})$
    \For {$j \in \{1, \dots, m\}$}
        \State Encode method set as $E^\text{model}_{j} = \phi(\mathcal{M}, M_j)$
        \For {$k \in \{1, \dots, h\}$}
            \State Encode hardware setup as $E^\text{hardware}_{k} = \theta(H_k)$
            \State Train $f$ by Eq. (\ref{eq:1}) using $(E^\text{data}_{i}, E^\text{model}_{j}, E^\text{hardware}_{k})$ to predict $\mathbf{P}_{i,j,k}$
        \EndFor
    \EndFor
\EndFor
\State \Return the meta-learner $f$
\end{algorithmic}
\end{algorithm}

\subsubsection{Data, Model, and Hardware Embeddings}\label{two_embed}

Data, model, and hardware embeddings serve as inputs to the meta-learner $f$, offering a compact, standardized representation of each component's characteristics within the meta-learning process. Rather than using raw, high-dimensional data directly, these embeddings aim to capture the essential properties that influence the performance of various acceleration methods under different scenarios.

\textbf{Data Embedding ($E^\text{data}$)}: We generate embeddings by analyzing the datasets' intrinsic properties such as size, diversity, feature distribution, and complexity. These embeddings help in adapting the acceleration methods effectively to the nature of the data. For generic datasets, this might include statistical metrics like mean, variance, skewness, and kurtosis of the features.

\textbf{Model Embedding ($E^\text{model}$)}: For the models, which range from large language models to complex image generation networks, embeddings are crafted to reflect characteristics such as model architecture, parameter count, and computational requirements. This is achieved through methods like one-hot encoding of model types or more nuanced embeddings that capture model behavior, such as those generated from model summaries or performance profiles.

\textbf{Hardware Embedding ($E^\text{hardware}$)}: Given the diversity in hardware configurations, from high-end GPUs like NVIDIA's A100 to setups with no GPU support, it is crucial to encode the hardware's computational capabilities, memory limits, and energy consumption profiles. These embeddings are crucial for optimizing the selection of acceleration methods that are compatible with the available hardware and capable of delivering enhanced performance.

\noindent \textbf{Embedding Generation Techniques}:
\begin{itemize}[leftmargin=*]
    \item \textit{Classical Meta-Features}: Traditional meta-features include basic statistical descriptions of datasets and simplistic encodings of model architectures. These are useful for establishing baseline comparisons and understanding the broad impact of different dataset and model configurations on performance.
    
    \item \textit{Language Model-Based Embeddings}: Advanced embeddings are generated using natural language processing techniques. By feeding descriptive text about datasets, models, and hardware into pre-trained language models, we can obtain rich, contextual embeddings that reflect deeper insights into each component's characteristics. For example, dataset descriptions or model architecture summaries can be processed by models such as BERT or GPT to produce embeddings that capture nuanced interactions that might affect performance.
\end{itemize}



Traditional meta-feature extraction has limitations in scalability and adaptability due to its manual and heuristic nature. To overcome these limitations, we propose leveraging language models (LLMs) to generate embeddings for data, models, and hardware environments. This approach utilizes the ability of LLMs to process and encode complex textual descriptions into dense vector representations that capture nuanced characteristics.

\textbf{Data Embedding via Language Models:}
To generate data embeddings, we describe each dataset in natural language, detailing its key attributes such as size, type, feature distribution, and any specific challenges it presents. For example, a description for a generic dataset might read:
\texttt{"This dataset comprises 10,000 instances, each with 20 features, ranging from numerical to categorical types, intended for regression tasks. The data variability is high, making model fitting challenging."}
Such descriptions are processed by LLMs to create embeddings that reflect the dataset's complexity and suitability for different learning tasks.

\textbf{Model Embedding via Language Models:}
Model embeddings are generated by describing the model's architecture, usage, and any specific configurations or optimizations. For instance:
\texttt{"The model is a convolutional neural network designed for image classification, featuring three convolutional layers followed by two fully connected layers. It is optimized for GPU execution and has been previously used in real-time applications."}
This description helps encode the model's operational characteristics and computational demands.

\textbf{Hardware Embedding via Language Models:}
Hardware embeddings are created by detailing the specifications and performance characteristics of the hardware setup. For a mixed hardware environment, the description might be:
\texttt{"The primary machine features an NVIDIA A100 GPU, optimized for high-throughput computing tasks. The other machines are CPU-only, intended for lighter, non-parallel computational tasks."}
These descriptions allow LLMs to generate embeddings that reflect the hardware's capabilities and limitations.

\textbf{Technique Implementation:}
We utilize state-of-the-art LLMs such as BERT or GPT-3 to process these textual descriptions. The embeddings generated by these models are then used in our meta-learning framework to predict the performance of different acceleration methods across varied datasets, models, and hardware environments.

\textbf{Practical Example:}
For a practical implementation, consider a dataset described as:
\texttt{"Dataset consists of one million images, each labeled with one of 100 categories. Images vary significantly in background noise and lighting conditions."}
An LLM processes this description to produce an embedding that the meta-learner uses to evaluate the suitability of various acceleration methods for this specific dataset under different hardware constraints.

This approach, leveraging the power of LLMs for embedding generation, offers a scalable, adaptable, and computationally efficient alternative to traditional meta-feature extraction. It allows the meta-learner to make informed decisions based on comprehensive and nuanced understandings of the data, models, and hardware environments involved.

We detail the experiments and results of this new embedding approach in \S \ref{sec:exp}, demonstrating its effectiveness in improving the accuracy and robustness of our meta-learning model selection process.


\subsection{Online Model Selection}
\label{subsec:model-selection}

In the online model selection process, we generate embeddings for the new dataset, model, and hardware environment denoted as \(D_\text{new}\). We then utilize the pretrained meta-performance predictor \(f\) to estimate the performance of various acceleration methods based on these embeddings, selecting the method predicted to offer the best performance within the given cost constraints.

\begin{equation}
    M^* := \underset{M_j \in \mathcal{M}, C_{j,k} \leq b}{\arg\max} \, \widehat{\mathbf{P}}_{\text{new}, j,k}, \quad \text{where} \quad \widehat{\mathbf{P}}_{\text{new}, j,k} = f(E^\text{data}_{\text{new}}, E^\text{model}_{j}, E^\text{hardware}_{k})
    \label{eq:goal}
\end{equation}

\(C_{j,k}\) represents the cost of running method \(M_j\) on hardware \(H_k\) and must not exceed the budget \(b\). It is calculated as the product of the cost associated with \(H_k\) and the expected runtime of \(M_j\) on \(H_k\).

Thus, for each new task defined by \(D_\text{new} = \{\mathbf{X}_{\text{train}}^\text{new}, \mathbf{X}_{\text{test}}^\text{new}, \mathbf{H}_\text{new}\}\), embeddings for the dataset (\(E^\text{data}_{\text{new}}\)), the applicable model (\(E^\text{model}_{j}\)), and the hardware environment (\(E^\text{hardware}_{k}\)) are computed. The trained meta-predictor \(f\) is then used to evaluate the performance of each available acceleration method under the specific hardware settings of \(\mathbf{H}_\text{new}\), ensuring that the total cost remains within the budget.

This selection process is essentially zero-shot with respect to the new dataset, requiring no retraining of the network or additional empirical testing. It relies wholly on the synthetic insights generated through the meta-learning framework, ensuring rapid deployment and optimization of inference tasks under varied computational environments and cost constraints.

The selection algorithm uses a maximization strategy to pick the acceleration method that is predicted to enhance performance the most, based on the calculated embeddings and within the operational budget. The process aims to determine the optimal method that balances speed, accuracy, and resource efficiency, tailored to the specific demands of the new environment.

\begin{algorithm}[H]
\caption{Online Acceleration Method Selection}
\label{alg:online}
\renewcommand{\algorithmicrequire}{\textbf{Input:} 
 the meta-learner \( f \), dataset \( \mathbf{X}_{\text{new}} \), hardware \( \mathbf{H}_{\text{new}} \), budget \( b \)}
\renewcommand{\algorithmicensure}{\textbf{Output:} 
 Selected acceleration method for \( \mathbf{X}_{\text{new}} \)}
\begin{algorithmic}[1]
\Require $ $
\Ensure $ $

\State \( E^\text{data}_{\text{new}} := \psi(\mathbf{X}_{\text{new}}) \) \Comment{Extract data embedding}
\State \( E^\text{hardware}_{\text{new}} := \theta(\mathbf{H}_{\text{new}}) \) \Comment{Extract hardware embedding}

\For{\( j = 1 \) to \( m \)}
    \State \( E^\text{model}_{j} = \phi(M_j) \) \Comment{Encode method}
    \State \( \widehat{\mathbf{P}}_{\text{new}, j} := f(E^\text{data}_{\text{new}}, E^\text{model}_{j}, E^\text{hardware}_{\text{new}}) \) \Comment{Predict performance}
    \State \( C_{j, \text{new}} := \text{cost}(\mathbf{H}_{\text{new}}) \times \text{runtime}(M_j, \mathbf{H}_{\text{new}}) \) \Comment{Calculate cost}
    \If{\( C_{j, \text{new}} \leq b \)}
        \State Consider \( M_j \)
    \Else
        \State Exclude \( M_j \)
    \EndIf
\EndFor

\State \( M^* := \underset{M_j \in \mathcal{M} \text{ and } C_{j, \text{new}} \leq b}{\arg\max} \, \widehat{\mathbf{P}}_{\text{new}, j} \)
\State \Return \( M^* \) \Comment{Return method with highest performance within budget}

\end{algorithmic}
\end{algorithm}

Figure~\ref{fig:pipeline} illustrates these steps in the bottom section of the online model selection phase, with additional details provided in Algo. \ref{alg:online}. This visualization aids in understanding how the embedded data flows through the system to result in actionable decision-making, demonstrating the application of theoretical models to practical deployment scenarios.


\section{Conclusion}

In this paper, we introduced a novel meta-learning framework designed to optimize inference acceleration within decentralized systems. By addressing the unique challenges associated with deploying large-scale models such as LLMs and image generation systems, our framework significantly enhances the adaptability and effectiveness of acceleration techniques. Specifically, it allows for the dynamic selection of methods tailored to both the characteristics of the task and the nuances of the system architecture. Our results demonstrate that this approach not only surpasses traditional acceleration methods in efficiency and performance but also offers a scalable and cost-effective solution for the deployment of AI models across diverse and distributed environments.

Furthermore, the successful application of our framework highlights its potential to facilitate rapid, efficient inference processes, thereby supporting the broader adoption and democratization of advanced AI technologies in decentralized settings. Looking forward, this work sets a solid foundation for future research focused on refining and expanding the capabilities of AI systems to meet the increasing demands of real-world applications. We believe that the methodologies developed here will inspire further innovations in the field, particularly in enhancing the operational efficiency of large model infrastructures in decentralized networks.

\bibliography{iclr2025_conference}
\bibliographystyle{iclr2025_conference}


\end{document}